\documentclass{article} 
\usepackage{iclr2021_conference,times}

\usepackage{amsmath,amsfonts,bm}

\def\eqref#1{equation~\ref{#1}}

\def\1{\bm{1}}

\DeclareMathAlphabet{\mathsfit}{\encodingdefault}{\sfdefault}{m}{sl}
\SetMathAlphabet{\mathsfit}{bold}{\encodingdefault}{\sfdefault}{bx}{n}

\usepackage{hyperref}
\renewcommand{\headrulewidth}{0pt}
\usepackage{booktabs}
\usepackage{url}
\usepackage{here}
\usepackage{multirow}
\usepackage{graphicx}

\title{Exploiting Topological Priors for Boosting Point Cloud Generation}

\author{Baiyuan Chen \\
College of Arts and Sciences \\
The University of Tokyo \\
Tokyo, Japan \\
\texttt{chenbaiyuan75@gmail.com}
}

\iclrfinalcopy 
\begin{document}

\maketitle

\begin{abstract}
    This paper presents an innovative enhancement to the Sphere as Prior Generative Adversarial Network (SP-GAN) model, a state-of-the-art GAN designed for point cloud generation. 
    A novel method is introduced for point cloud generation that elevates the structural integrity and overall quality of the generated point clouds by incorporating topological priors into the training process of the generator. 
    Specifically, this work utilizes the K-means algorithm to segment a point cloud from the repository into clusters and extract centroids, which are then used as priors in the generation process of the SP-GAN. 
    Furthermore, the discriminator component of the SP-GAN utilizes the identical point cloud that contributed the centroids, ensuring a coherent and consistent learning environment. 
    This strategic use of centroids as intuitive guides not only boosts the efficiency of global feature learning but also substantially improves the structural coherence and fidelity of the generated point clouds. 
    By applying the K-means algorithm to generate centroids as the prior, the work intuitively and experimentally demonstrates that such a prior enhances the quality of generated point clouds.
\end{abstract}

\section{Introduction}

3D point cloud generation is an important field in computer vision and is utilized broadly in diverse fields such as VR immersive environment creation [1]. 
It offers comprehensive geometric details regarding the form and surface attributes of physical items and environments. To generate point clouds with high quality, one useful approach is to capture the local and global features of a point cloud. 
While numerous studies have focused on capturing these features during the generation process, the integration of topological priors in the generation process remains relatively unexplored. 
A topological prior can serve as a hint for global feature generation, hence improving the training efficiency and generating high-quality point clouds.

This study concatenates a topological prior into the prior latent matrix of the Sphere as Prior Generative Adversarial Network (SP-GAN model) [2]. 
SP-GAN is a point cloud generation model that is based on a generative adversarial network (GAN). The authors utilize a sphere as the initial state and adapt the latent vectors associated with this sphere to enhance the generation and manipulation of point clouds. 
To implement the approach, this work selects centroids from a point cloud within the repository, adjusts their dimensions to match those of the initial sphere, and then concatenates these modified centroids with the SP-GAN's prior latent matrix. 
This process introduces a topological prior into the generation phase. This paper intuitively and experimentally showed that with such a prior, the SP-GAN can generate point clouds with higher quality.

\section{Related Work}

Various approaches have been explored in 3D shape generation. Volumetric Methods represent 3D shapes in a volumetric grid, similar to pixels in 2D images but extended into three dimensions [3]. Implicit surface methods represent 3D shapes from latent space [4]. 
Another approach is based on point cloud generation. Point cloud generation also includes various approaches such as Transformer-based models, 3D deep learning, etc.

\subsection{Transformer-based models}

Transformer-based models utilize the Transformer to generate point clouds. Even though the Transformer is originally designed for natural language processing, it is being widely used in other fields. 
By leveraging its architecture, it has been proven effective in point cloud processing, and because of its strength in global feature learning, it has been applied to a wide range of point cloud processing tasks such as segmentation, classification, and generation [5]. 
Its attention mechanism has also proved to be effective for point cloud processing tasks [6].

\subsection{3D Deep Learning}

3D Deep Learning has progressed tremendously in recent years. Many 3D deep learning models are proposed [7]. As one of the basic deep neural network structures, GAN has become particularly influential in the domain of point cloud generation. 
Tree-GAN generates tree-like structures through a hierarchical process that mirrors natural tree growth, focusing on the branching complexity and diversity of the generated point clouds [8]. PointInverter inverts point cloud generation processes, which reconstruct precise 3D shapes from their latent representations [9]. 
SP-GAN generates point clouds by emphasizing spatial relationships within point clouds, which effectively balance the local and global features [2]. This work builds upon the foundation laid by SP-GAN, integrating a topological prior into the training process of the generator, enhancing the quality of the generated point clouds.

\section{Methodologies}

In this section, the proposed point cloud generation model is introduced. As shown in Fig. 1, the model primarily utilizes the SP-GAN architecture but incorporates a unique enhancement: the inclusion of a prior during the generation process, derived using the K-means algorithm. 
In Sections 3.1 to 3.3, this work elaborates on the model's framework, detailing its structure and the integration of the prior. In Section 3.4, this work presents details of training and evaluation.

\subsection{Overview}

This work aims to train the generator of SP-GAN to generate high-quality point clouds, which have dense correspondence with the initial global state (sphere) and the reference point cloud. 
As shown in Fig. 1, the framework of the model consists of three parts: a prior latent matrix, a generator $\mathcal{G}$, and a discriminator $\mathcal{D}$. The process begins with the construction of a prior latent matrix, which is subsequently fed into the generator. 
The generator then utilizes this input to generate a point cloud $\mathcal{P}$. Finally, this generated point cloud $\mathcal{P}$ or a reference point cloud $\hat{\mathcal{P}}$ from the repository is presented to the discriminator. 
The discriminator evaluates the input and assigns two types of scores: per-point scores and a per-shape score, to decide whether the point cloud is an original from the repository or a generated one.

\begin{figure}[H]
    \centering
    \includegraphics[scale=0.3]{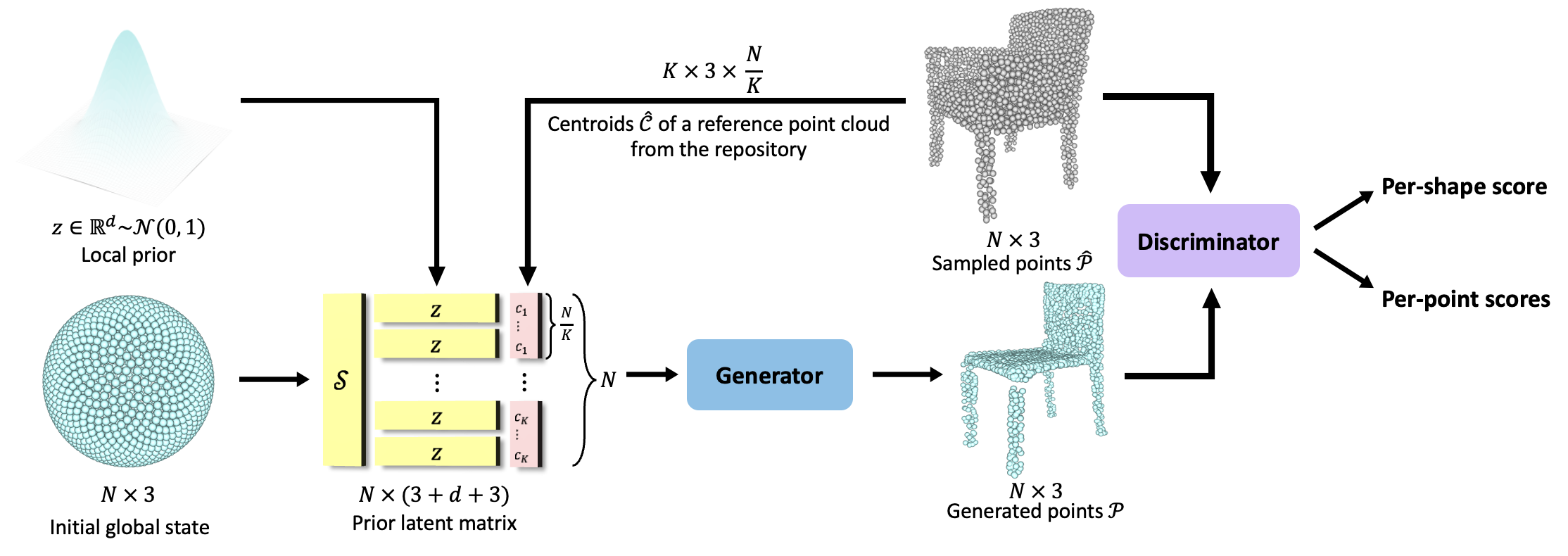}
    \caption{The framework of the proposed modified SP-GAN model.}
    \label{fig1}
\end{figure}

\subsection{Prior Latent Matrix}

As shown in Fig. 1, the prior latent matrix consists of three components: the initial global state represented by a point cloud sphere, a set of local priors, and the centroids of a reference point cloud from the repository. 
The initial two components align with the standard SP-GAN setup: a uniformly distributed point cloud sphere $\mathcal{P}$ with 2048 points, and the concatenation of $N$ vectors $z\in\mathbb{R}^d$. 
Here, the point cloud sphere represents the initial global state, while the vector $z\in\mathbb{R}^d$ represents the latent vector of each point on the sphere, with each element drawn independently from a standard normal distribution. 
The last part of the prior latent matrix presents a topological prior. Such prior consists of the centroids of a point cloud from the repository, which is also the one fed into the discriminator. 
To establish this, this work uses the K-means algorithm to divide the point cloud into K clusters, where K is a factor of 2048. Then, the centroid $c_i\,\,(i=1,\cdots,K)$ is taken of each cluster, replicate each centroid $\frac{N}{K}$ times and concatenate them to forge a matrix $\hat{\mathcal{C}}\in \mathbb{R}^{N\times 3}$. 
Instead of forming a matrix of all centroids first and then replicating it $\frac{N}{K}$ times, this work empirically finds that duplicating centroids first and then operating concatenation can generate point clouds with higher quality, 
which means the shapes of generated point clouds are more complete in shape and exhibit a greater correspondence with the reference point clouds in the repository. 

\begin{figure}[H]
    \centering
    \includegraphics[scale=0.4]{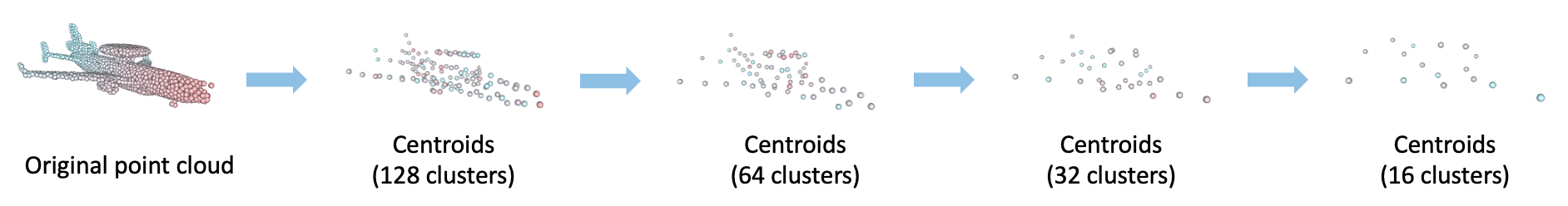}
    \caption{Demonstration of calculating centroids of a point cloud.}
    \label{fig2}
\end{figure}

\subsection{Generator and Discriminator}

The architecture of the proposed model's generator and discriminator remains largely consistent with those outlined in the original SP-GAN framework, except for the input dimension of the feature embedding module, which is $N\times(3+d+3)$ instead of $N\times(3+d)$ because of the inclusion of the topological prior (Fig. 1). 
This work sets $N=2048,\,\,d=128$, and the augmented prior latent matrix is fed into the feature embedding module. After processing by the generator, the local and global features of the prior latent matrix are extracted, with style features embedded. This feature extraction leads to the generation of a point cloud $\mathcal{P}$.

When the discriminator evaluates an input point cloud, it extracts point features and outputs two distinct scores based on these features: per-point scores and a per-shape score. The discriminator can then utilize the per-shape score to stabilize overall differences within the input point cloud, while the per-point scores are for local variation regulation. 
In this way, this dual-scoring mechanism allows the discriminator to outperform traditional discriminators by providing regulation of both global and local variations in point clouds.

\subsection{Training Details and Evaluation of Point Cloud Generation}

During training, the same loss function setup as the SP-GAN paper is used, which is based on the least square loss method from [10]. The loss functions of the generator $\mathcal{L}_G$ and the discriminator $\mathcal{L}_D$ are comprised of two components: shape loss and point loss. The functions are shown below, where the main difference between $\mathcal{L}_G$ and $\mathcal{L}_D$ is that $\mathcal{L}_G$ includes an adjustment for point loss to ensure the point loss and the shape loss are in the same scale:
\begin{equation}
    \mathcal{L}_G = \frac{1}{2} \left[ D(\mathcal{P}) - 1 \right]^2 + \beta \frac{1}{2N} \sum_{i=1}^N \left[ D(p_i) - 1 \right]^2
\end{equation}

\begin{equation}
    \mathcal{L}_D = \mathcal{L}_D^{\text{shape}} + \mathcal{L}_D^{\text{point}}
\end{equation}

\begin{equation}
    \mathcal{L}_D^{\text{shape}} = \frac{1}{2} \left( \left[ D(\mathcal{P}) - 1 \right]^2 + \left[ D(\hat{\mathcal{P}}) - 1 \right]^2 \right)
\end{equation}

\begin{equation}
    \mathcal{L}_D^{\text{point}} = \frac{1}{2N} \sum_{i=1}^N \left( \left[ D(p_i) - 1 \right]^2 + \left[ D(\hat{p}_i) - 1 \right]^2 \right)
\end{equation}
where $\beta$ is a parameter for scale balance; $p_i$ and $\hat{p}_i$ are the i-th point in $\mathcal{P}$ and $\hat{\mathcal{P}}$ , respectively.

Since the topological prior is concatenated with the prior latent matrix in the training process, point cloud generation in the evaluation phase should be different from the SP-GAN paper. 
To keep the input dimension consistent with the trained generator, this work concatenates a unit sphere $\mathcal{S}$, $N$ latent vectors $z$, and the same unit sphere $\mathcal{S}$ again in the evaluation phase. This ensures that the input dimension to the generator during evaluation matches what it was trained with.

The proposed model is trained on airplane, chair, car, and guitar categories from the ShapeNet dataset for 100 epochs. After that, the trained generator is used to create 1000 samples for testing. To evaluate the quality of these generated point clouds, two metrics are utilized: the Fréchet Point Cloud Distance (FPD) and the Jensen-Shannon Divergence (JSD) [11,12]. 
Similar to the Frechet Inception Distance (FID) [13], the FPD measures the 2-Wasserstein distance between genuine and synthetic Gaussian distributions within the model-derived feature spaces. On the other hand, the JSD represents a symmetric and smoothed adaptation of the Kullback-Leibler divergence [14], enabling the assessment of similarity between two probability distributions [12].

\section{Results}

\subsection{Quantitative Results}

This work utilizes centroids from 16, 32, 64, and 128 clusters, and also the original point cloud as the topological prior. When a reference point cloud is concatenated with the prior latent matrix, it can be regarded as adding centroids of 2048 clusters. The overall results are shown in Table 1. The unit of FPD is $10^{-3}$. 
From this table, it's clear that while the original SP-GAN performs best in terms of the FPD for chairs and the JSD for cars, incorporating a topological prior generally improves the quality of generated point clouds across most scenarios compared to the baseline SP-GAN. 
A possible explanation for the modified SP-GAN's lower performance with cars might be the high similarity in topological structures among car point clouds in the dataset, as shown in Fig. 3. This similarity could render the topological prior less effective and more redundant. However, the reason for the chair category's results is not as clear.

\begin{table}[h]
    \caption{Performance Comparison: Original SP-GAN vs. Topological Prior-Enhanced SP-GAN \\\qquad\qquad\qquad\qquad\qquad\qquad\quad(Airplane, Chair, Car, Guitar).}
    \label{sample-table}
    \begin{center}
    \begin{tabular}{l l l l l l l l l}
    \multirow{2}{*}{\bf Categories}  &\multirow{2}{*}{\bf Metrics} &\multirow{2}{*}{\bf Vanilla} &\multicolumn{5}{c}{\bf Number of Centroids Used} \\
    \cline{4-8}
    &&& \multicolumn{1}{c}{\bf 2048} & \multicolumn{1}{c}{\bf 128} & \multicolumn{1}{c}{\bf 64} & \multicolumn{1}{c}{\bf 32} & \multicolumn{1}{c}{\bf 16} \\
    \hline
    \multirow{2}{*}{Airplane} &FPD($\downarrow$)&10.54&10.87&\textbf{4.23}&8.41&13.60&5.35 \\
             &JSD($\downarrow$)&0.9&0.86&0.85&\textbf{0.84}&0.87&0.89 \\
    \cline{2-8}
    \multirow{2}{*}{Chair} &FPD($\downarrow$)&\textbf{7.90}&11.79&12.11&12.47&13.82&11.23 \\
             &JSD($\downarrow$)&0.93&0.90&0.88&0.89&0.89&\textbf{0.85} \\
    \cline{2-8}
    \multirow{2}{*}{Car} &FPD($\downarrow$)&5.46&4.57&3.46&8.15&7.40&\textbf{3.34} \\
             &JSD($\downarrow$)&\textbf{0.95}&0.98&0.99&0.96&0.98&0.99 \\
    \cline{2-8}
    \multirow{2}{*}{Guitar} &FPD($\downarrow$)&4.00&8.49&3.77&\textbf{3.59}&5.92&4.11 \\
             &JSD($\downarrow$)&0.90&0.98&0.98&0.94&\textbf{0.87}&0.96 \\
    \hline
    \end{tabular}
    \end{center}
\end{table}

\subsection{Qualitative Results}

Fig. 3 shows generated samples of the modified SP-GANs. The airplane samples are generated by the model with 128 centroids, the chair samples with 16 centroids, the car samples with 2048 centroids (original point cloud), and the guitar samples with 64 centroids. From the result, it could be observed that even with just 100 epochs of training, the models can still produce high-quality point clouds.

\begin{figure}
    \centering
    \includegraphics[scale=0.4]{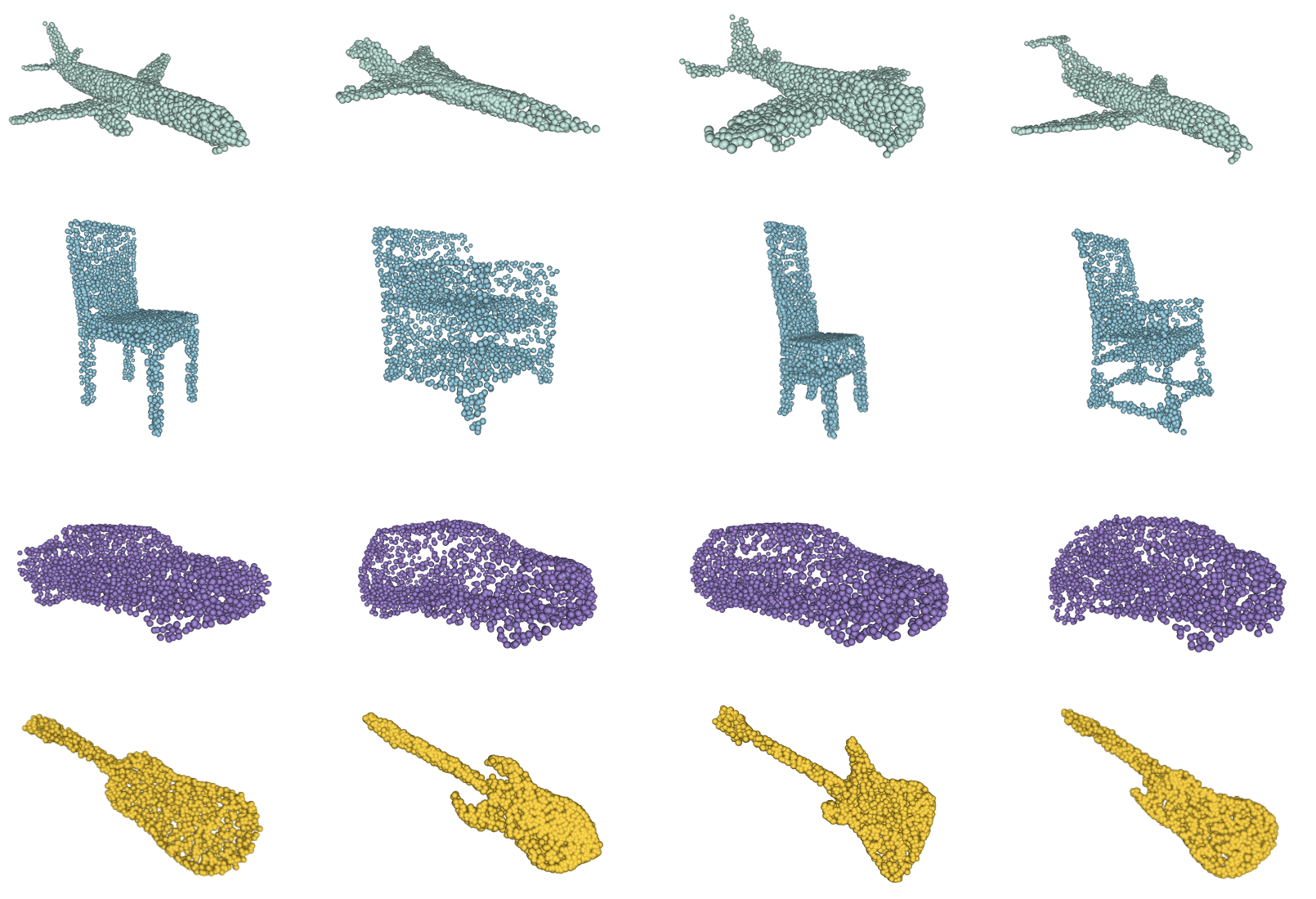}
    \caption{Generated samples of airplane, chair, car, and guitar.}
    \label{fig3}
\end{figure}

\section{Discussion}

This work evaluated the modified SP-GANs with 4 categories of point clouds from the ShapeNet dataset and found that these models, on average, generated point clouds with lower FPD and JSD compared to the original SP-GAN. Furthermore, when the standard SP-GAN is trained on the airplane category for 300 epochs, an FPD of $8.52\times10^{-3}$ and a JSD of 0.85 are achieved. 
Comparing these figures to the results from the SP-GAN with a 64-centroid prior, it's evident that the modified SP-GAN can generate better quality point clouds, potentially reducing the need for a large number of training epochs.

Despite the overall success, the performance in generating chair and car point clouds was not as distinct. This work noticed that car point clouds tend to have more similar shapes compared to other categories, suggesting that the topological prior works best with more unique shapes. Also, while the modified SP-GAN generally outperforms the original, the improvements can sometimes be minor, and improvements in JSD and FPD don't always occur together. 

Additionally, this work has not yet tested the topological prior approach on other models, so its wider applicability remains unconfirmed. However, the method introduces a new way of generating point clouds, which might be capable of implementing on other point cloud processing models by modifying the structure of the topological prior, leading to higher performance.

\section{Conclusion}

This study introduces a straightforward yet effective strategy for enhancing point cloud generation. This method involves creating a topological prior by concatenating centroids from a reference point cloud, obtained through the K-means algorithm. This study incorporates the prior within the SP-GAN structure, showcasing its ability to enhance the quality of the resulting point clouds. This approach, while simple, represents a novel contribution to the field of point cloud generation.

For future work, it is worth implementing such prior on other point cloud generation models to expand its generality. Also, the quality of generated point clouds still needs to be improved, which involves adjusting the structure of the topological prior. Furthermore, the application of the topological prior can be extended to additional point cloud processing tasks like segmentation and classification to evaluate its efficacy.

\bibliography{iclr2021_conference}
\bibliographystyle{iclr2021_conference}

[1]	Nießner, Matthias, Michael Zollhöfer, Shahram Izadi, and Marc Stamminger. Real-time 3D reconstruction at scale using voxel hashing. ACM Transactions on Graphics, 2013, 32(6): 1-11.\\\quad\\\quad
[2]	Li, Ruihui, Xianzhi Li, Ka-Hei Hui, and Chi-Wing Fu. SP-GAN: Sphere-guided 3D shape generation and manipulation. ACM Transactions on Graphics, 2021, 40(4): 1-12.\\\quad\\\quad
[3]	Wu, Chengzhi, Junwei Zheng, Julius Pfrommer, and Jürgen Beyerer. Attention-based Part Assembly for 3D Volumetric Shape Modeling. In Proceedings of the IEEE/CVF Conference on Computer Vision and Pattern Recognition, 2023: 2716-2725.\\\quad\\\quad
[4]	Jangid, Devendra K., Neal R. Brodnik, Amil Khan, McLean P. Echlin, Tresa M. Pollock, Sam Daly, and B. S. Manjunath. 3DMaterialGAN: Learning 3D Shape Representation from Latent Space for Materials Science Applications. ArXiv Preprint, 2020: 2007.13887.\\\quad\\\quad
[5]	Lu, Dening, Qian Xie, Mingqiang Wei, Kyle Gao, Linlin Xu, and Jonathan Li. Transformers in 3d point clouds: A survey. ArXiv Preprint, 2022: 2205.07417.\\\quad\\\quad
[6]	Wang, Lei, Yuchun Huang, Yaolin Hou, Shenman Zhang, and Jie Shan. Graph attention convolution for point cloud semantic segmentation. In Proceedings of the IEEE/CVF conference on computer vision and pattern recognition, 2019: 10296-10305.\\\quad\\\quad
[7]	Ioannidou, Anastasia, Elisavet Chatzilari, Spiros Nikolopoulos, and Ioannis Kompatsiaris. Deep learning advances in computer vision with 3d data: A survey. ACM computing surveys, 2017, 50(2): 1-38.\\\quad\\\quad
[8]	Shu, Dong Wook, Sung Woo Park, and Junseok Kwon. 3d point cloud generative adversarial network based on tree structured graph convolutions. In Proceedings of the IEEE/CVF international conference on computer vision, 2019: 3859-3868.\\\quad\\\quad
[9]	Kim, Jaeyeon, Binh-Son Hua, Thanh Nguyen, and Sai-Kit Yeung. Pointinverter: Point cloud reconstruction and editing via a generative model with shape priors. In Proceedings of the IEEE/CVF Winter Conference on Applications of Computer Vision, 2023: 592-601.\\\quad\\\quad
[10]	Mao, Xudong, Qing Li, Haoran Xie, Raymond YK Lau, Zhen Wang, and Stephen Paul Smolley. Least squares generative adversarial networks. In Proceedings of the IEEE international conference on computer vision, 2017: 2794-2802.\\\quad\\\quad
[11]	Shu, Dong Wook, Sung Woo Park, and Junseok Kwon. 3d point cloud generative adversarial network based on tree structured graph convolutions. In Proceedings of the IEEE/CVF international conference on computer vision, 2019: 3859-3868.\\\quad\\\quad
[12]	Achlioptas, Panos, Olga Diamanti, Ioannis Mitliagkas, and Leonidas Guibas. Learning representations and generative models for 3d point clouds." In International conference on machine learning, 2018: 40-49.\\\quad\\\quad
[13]	Heusel, Martin, Hubert Ramsauer, Thomas Unterthiner, Bernhard Nessler, and Sepp Hochreiter. Gans trained by a two time-scale update rule converge to a local nash equilibrium. Advances in neural information processing systems 2017, 30: 1-12.\\\quad\\\quad
[14]	Kullback, Solomon, and Richard A. Leibler. On information and sufficiency. The annals of mathematical statistics, 1951, 22(1): 79-86.

\end{document}